\documentclass[preprint,12pt,authoryear]{elsarticle}

\usepackage{amssymb}
\usepackage{amsmath}

\usepackage{natbib}
\usepackage{bibentry}
\usepackage{tikz}
\usepackage{xcolor}
\usepackage{enumitem}
\usepackage{graphicx}
\usepackage{booktabs}
\usepackage{siunitx}
\usepackage{multirow}
\usepackage{algorithm}
\usepackage{algorithmic}
\usepackage[table]{xcolor}
\definecolor{lightviolet}{RGB}{200, 185, 227}

\usepackage{pifont}

\usepackage{booktabs, multirow, tabularx}
\usepackage{siunitx} 
\usepackage{adjustbox}

\journal{Nuclear Physics B}

\begin{document}

\begin{frontmatter}



\title{ASTIF: Adaptive Semantic-Temporal Integration for Cryptocurrency Price Forecasting} 


\author[SWUFE]{Hafiz Saif Ur Rehman}
\ead{saif_@msn.com}

\author[SWUFE]{Ling Liu\corref{cor1}}
\ead{lingliu@swufe.edu.cn}

\author[SWJTU]{Kaleem Ullah Qasim}
\ead{kaleemullah@swjtu.edu.cn}

\cortext[cor1]{Corresponding author}

\affiliation[SWUFE]{organization={Southwestern University of Finance and Economics},
            addressline={No.555 Liutai Avenue, Wenjiang District},
            city={Chengdu},
            state={Sichuan},
            postcode={611130},
            country={China}}

\affiliation[SWJTU]{organization={Southwest Jiaotong University},
            addressline={No.111 North Section 1, Second Ring Road},
            city={Chengdu},
            state={Sichuan},
            postcode={610031},
            country={China}}
\begin{abstract}
Financial time series forecasting is fundamentally an information fusion challenge, yet most existing models rely on static architectures that struggle to integrate heterogeneous knowledge sources or adjust to rapid regime shifts. Conventional approaches, relying exclusively on historical price sequences, often neglect the semantic drivers of volatility such as policy uncertainty and market narratives. To address these limitations, we propose the ASTIF (Adaptive Semantic-Temporal Integration for Cryptocurrency Price Forecasting), a hybrid intelligent system that adapts its forecasting strategy in real time through confidence-based meta-learning. The framework integrates three complementary components. A dual-channel Small Language Model using MirrorPrompt extracts semantic market cues alongside numerical trends. A hybrid LSTM–Random Forest model captures sequential temporal dependencies. A confidence-aware meta-learner functions as an adaptive inference layer, modulating each predictor’s contribution based on its real-time uncertainty.
Experimental evaluation on a diverse dataset of AI-focused cryptocurrencies and major technology stocks from 2020 to 2024 shows that ASTIF outperforms leading deep learning and Transformer baselines (e.g., Informer, TFT). The ablation studies further confirm the critical role of the adaptive meta-learning mechanism, which successfully mitigates risk by shifting reliance between semantic and temporal channels during market turbulence. The research contributes a scalable, knowledge-based solution for fusing quantitative and qualitative data in non-stationary environments.
\end{abstract}


\end{frontmatter}


\section{Introduction}
\label{sec:introduction}
Cryptocurrency markets have become an increasingly important component of global finance, with daily trading volumes exceeding \$2 trillion as of October 2025 \citep{chokor2021effect, statista2025crypto}. Unlike traditional financial assets, cryptocurrencies operate within a market structure characterized by pronounced price volatility, uninterrupted trading activity, and heightened exposure to heterogeneous external drivers \citep{watorenek2023crypto, ghergina2025volatility}. These drivers are not limited to macroeconomic conditions but also encompass geopolitical developments, regulatory interventions, and shifts in investor sentiment amplified through social media channels \citep{deleon2022bitcoin}. Taken together, such features give rise to a market environment in which price formation is highly unstable and difficult to model. As a result, conventional forecasting approaches frequently prove inadequate for capturing the non-linear and rapidly evolving dynamics observed in cryptocurrency prices \citep{kehinde2025helformer}. Accurate price forecasting in this setting is thus practically significant for institutional investors managing risk, for financial institutions developing digital asset strategies, and for regulators focused on preserving market stability.
 
Recent work on cryptocurrency forecasting research faces a fundamental paradox. Markets exhibit regime-dependent dynamics where prediction strategies must adapt to evolving conditions \citep{sharma2022performance, petropoulos2022forecasting}, yet dominant methodological paradigms employ static architectures with fixed integration strategies. Statistical approaches prioritize interpretability through linear relationships \citep{khedr2021cryptocurrency}, while deep learning architectures emphasize temporal dependency modeling \citep{fischer2018deep, seabe2023forecasting}. Building on these foundations, recent hybrid frameworks pursue multimodal integration by combining neural networks with various data sources, including social sentiment, blockchain metrics, and volatility spillovers \citep{guo2024mfb, deleon2022bitcoin, kim2022deep, fu2025quantile}. Decomposition-aided approaches have demonstrated effectiveness in cryptocurrency forecasting by extracting temporal modes from volatile price series \citep{mizdrakovic2024forecasting}. Despite their architectural heterogeneity, these paradigms converge on a common limitation. Whether employing LSTM-CNN architectures for multi-scale feature extraction \citep{livieris2021advanced}, graph networks for cryptocurrency interrelations \citep{zhong2023lstm}, or attention mechanisms for dynamic relationship modeling \citep{choi2025foretell}, existing frameworks maintain fixed model weights and predetermined integration rules. When the market microstructure changes around regulatory announcements or macroeconomic events, these systems have limited ability to rebalance their prediction strategy \citep{xie2023wall}. This architectural rigidity manifests itself in systematic performance degradation during regime transitions \citep{sharma2022performance}. The core methodological gap thus becomes clear: current research lacks adaptive mechanisms that dynamically select between heterogeneous prediction paradigms based on real-time confidence evaluation and market context, particularly for integrating semantic understanding of policy uncertainty with temporal pattern recognition.

To address the gap, we propose the Adaptive Semantic-Temporal Integration Framework (ASTIF). A hybrid system that enables adaptive model selection for cryptocurrency price forecasting. The framework comprises three integrated components working in concert to optimize prediction accuracy. First, MirrorPrompt is a dual-channel small language model (SLM) architecture that processes numerical price sequences and semantic market indices through separate computational pathways. The design enables simultaneous reasoning across both data modalities. Second, an LSTM-based temporal predictor captures long-range dependencies in historical price movements, providing a strong baseline grounded in temporal patterns. Third, a meta-learning module evaluates the confidence in prediction from both channels and performs adaptive model selection. Dynamically weights the semantic and temporal predictions based on their respective confidence scores. To validate the reliability of the framework across diverse market regimes, we evaluated ASTIF on a multi-asset dataset of AI-focused cryptocurrencies and traditional technology benchmarks covering the 2020–2024 period.

The contributions of this study are threefold. Methodologically, a confidence-based meta-learning framework enables dynamic selection among heterogeneous predictors, overcoming the limitations of static architectures during market regime shifts. Architecturally, a dual-channel architecture integrates semantic reasoning via Small Language Models and temporal patterns through LSTM predictors, processing market indices, such as policy uncertainty and cross-asset correlations, along separate pathways to preserve contextual information. Operationally, a confidence-weighted ensemble uses real-time uncertainty estimates to adjust strategies adaptively and mitigate risks under extreme market volatility.

The remainder of this paper is organized as follows: Section~\ref{sec:related_work} describes literature review; Section~\ref{sec:methodology} formalizes the forecasting problem and describes the ASTIF framework; Section~\ref{sec:experiments} presents the experimental setup; Section~\ref{sec:results} reports experimental results and ablation studies; and Section~\ref{sec:conclusion} concludes the paper, limitations and future work.

\section{Literature Review}
\label{sec:related_work}
Financial time series forecasting, particularly for volatile cryptocurrency markets, presents fundamental challenges that existing approaches have yet to adequately address. Accurate cryptocurrency price prediction has significant implications for portfolio optimization, risk management, and algorithmic trading strategies in increasingly digitalized financial markets. This section examines three critical challenges: the limitations of static architectures in non-stationary market environments, the superficial integration of semantic information in current hybrid models, and the absence of adaptive mechanisms for heterogeneous predictor selection.

\subsection{Challenges in Modeling Non-Stationary Market Dynamics}
\label{subsec:nonstationarity}
Cryptocurrency markets exhibit extreme volatility, rapid regime shifts, and structural breaks that challenge conventional forecasting architectures. Classical econometric models such as ARIMA, GARCH, and Vector Autoregression (VAR) assume market stationarity and linear relationships \citep{nakano2018bitcoin}. These assumptions render them ineffective for capturing regime shifts, asymmetric volatility, and abrupt price movements characteristic of cryptocurrency markets. Deep learning architectures, particularly LSTM networks \citep{seabe2023forecasting, sciencedirect2024hybrid}, address temporal dependencies more effectively than statistical methods \citep{CHEN2023187}. However, LSTM-based approaches suffer from critical deficiencies. Such networks process only numerical price sequences without interpreting contextual information such as policy uncertainty. Recurrent neural architectures also lack uncertainty quantification mechanisms for prediction reliability and employ static configurations that cannot adapt to regime changes. Recent Transformer-based models have demonstrated superior performance in traditional time series forecasting \citep{zhou2021informer, Lai2017ModelingLA}, yet their application to cryptocurrency markets reveals fundamental limitations. The Informer Transformer struggles with high-frequency fluctuations in crypto data \citep{nie2023patchtst}, while the Temporal Fusion Transformer (TFT) faces challenges in non-stationary environments \citep{lim2021temporal}, where direct application results in error amplification during volatile periods. Hybrid architectures combining BiLSTM with modified Transformers and temporal convolutional networks have shown promise in stock prediction by capturing multi-scale temporal dependencies \citep{tian2025bimt}, yet these approaches still employ fixed integration weights and cannot dynamically rebalance prediction strategies across market regimes. 

Critically, conventional forecasting architectures, whether statistical, recurrent, or attention-based, share a common weakness: static model configurations with fixed parameters that cannot dynamically adjust when market conditions shift. Cryptocurrency markets experience regime transitions triggered by regulatory announcements, technological developments, and macroeconomic events that fundamentally alter price dynamics within hours. Existing architectures lack the adaptive capacity to rebalance their prediction strategies in real-time, resulting in degraded performance during market regime transitions. These models also operate exclusively on numerical price data, ignoring semantic market context embedded in policy uncertainty indices, cross-asset correlations, and qualitative market signals. Above limitation is particularly problematic because Transformers like TFT, despite their sophisticated attention mechanisms, cannot interpret the semantic meaning of regulatory announcements or policy shifts. Attention-based architectures process only historical numerical patterns. This gap motivates the need for frameworks that combine adaptive model selection with semantic understanding of structured market indices.

\subsection{The Role of Semantic Market Data in Financial Prediction}
\label{subsec:semantic_data}
Recent research has explored integrating textual information into financial forecasting through Natural Language Processing techniques \citep{kraaijeveld2020predictive}. \citep{kim2023deep} fine-tuned RoBERTa transformers on cryptocurrency news, while \citep{frontiers2025short} employed BERT and GloVe embeddings for analyzing news headlines in short-term forecasting. Social media sentiment, particularly Twitter data, has been shown to influence cryptocurrency price movements \citep{critien2022bitcoin}. Hybrid methodologies combine these NLP-derived features with temporal models: \citep{jahanbin2025cryptocurrency} integrated wavelet transforms with LSTM, \citep{balijepalli2024prediction} employed ensemble methods using gradient-boosted trees and random forests, and \citep{mahdi2025crypto} explored a deep learning hybrid model that integrates attention Transformer and GRU architectures. \citep{farimani2022investigating} demonstrated that combining FinBERT-based sentiment analysis of financial news with technical indicators improves forex and cryptocurrency price prediction, though their approach concatenates features rather than processing them through separate reasoning channels. However, these approaches exhibit a fundamental limitation in extracting sentiment as scalar features appended to price data rather than processing semantic information through architectures that employ contextual reasoning. This approach treats complex news narratives, regulatory announcements, and policy uncertainty as single numerical scores, causing structural meaning and contextual relationships that influence market dynamics to be lost.

This shallow integration undermines the potential of semantic data in financial forecasting. Unlike simple sentiment scoring that reduces multidimensional policy uncertainty to a scalar value, effective semantic processing requires architectures that can reason about contextual implications of structured market indices. These indices include Global Economic Policy Uncertainty (GEPU), Geopolitical Risk Deviation (GPRD), cross-asset correlations, and volatility patterns. The emergence of Small Language Models offers computational efficiency while maintaining reasoning capabilities \citep{wan2024efficient, brown2020language}. However, their application to financial forecasting through structured prompt engineering remains underexplored. Recent financial LLMs such as FinRipple \citep{finripple2025} have demonstrated potential for aligning language models with market data, but these approaches lack integration with temporal neural networks and do not provide uncertainty-aware fusion mechanisms. This creates reliability concerns in production systems where overconfident predictions lead to significant losses. The critical gap lies in establishing symmetric dual-channel architectures that process semantic market indices and numerical price sequences with equal representational capacity. Such architectures would enable models to disentangle qualitative policy signals from quantitative temporal patterns.

\subsection{Adaptive Learning Paradigms and Meta-Learning}
\label{subsec:meta_learning}
Meta-learning has emerged as a paradigm for adaptive forecasting, enabling dynamic strategy selection based on market conditions \citep{finn2017model}. \citep{tailored2025meta} proposed meta-learning frameworks for short time series addressing cold-start problems in new markets. \citep{evaluationfree2025meta} introduced evaluation-free model selection through learned policies, while \citep{instancebased2023meta} developed instance-based approaches for multi-step forecasting. \citep{vaiciukynas2023meta} demonstrated few-shot learning for cryptocurrency prediction under data scarcity. These advances represent progress toward adaptive architectures, yet existing meta-learning frameworks focus exclusively on selecting among homogeneous predictors, such as choosing the best-performing LSTM configuration or optimal hyperparameters for similar model families.

This homogeneity constraint represents a critical limitation. Current meta-learning approaches do not address integrating heterogeneous predictors with fundamentally different processing paradigms. Frameworks that dynamically switch between semantic reasoning channels and temporal pattern channels based on prediction confidence remain absent from the literature. Semantic reasoning channels employ language models interpreting structured market indices, while temporal pattern channels use LSTM networks capturing numerical dependencies. Existing approaches also lack mechanisms for uncertainty quantification from heterogeneous sources and confidence-calibrated ensembling that adaptively weights predictions based on real-time reliability estimates. This gap is particularly problematic because cryptocurrency markets exhibit varying dependencies on semantic versus temporal signals across different volatility regimes. During stable periods, temporal patterns may dominate. Conversely, during policy-driven events, semantic understanding becomes critical. Without adaptive cross-modal selection mechanisms, forecasting architectures cannot optimally exploit the complementary strengths of heterogeneous predictors.

\subsection{Research Gaps and Objectives}
\label{subsec:research_gaps}
Despite significant advancements across these approaches, three critical gaps persist in cryptocurrency forecasting research. The most prominent limitation is architectural rigidity, where existing models, whether statistical, deep learning, or Transformer-based, rely on static configurations that cannot dynamically adjust prediction strategies during market regime shifts. Compounding this issue is the prevalence of semantic asymmetry, a tendency to reduce complex policy contexts to simple scalar sentiment scores, thereby stripping away the structural meaning required to interpret price dynamics. Finally, the literature reveals a significant lack of cross-modal adaptation, specifically regarding the absence of frameworks capable of weighing the confidence of semantic reasoning against temporal pattern recognition based on real-time uncertainty.

Addressing these gaps requires a unified framework that integrates a dual-channel architecture for the symmetric processing of semantic and numerical data, while simultaneously employing confidence-based meta-learning to dynamically arbitrate between these heterogeneous predictors. To assess the efficacy of the proposed ASTIF framework, this study investigates the following three research questions:

\textbf{RQ1. Comparative Performance against Baselines.} To what extent does the ASTIF framework demonstrate superior forecasting accuracy and stability compared to traditional machine learning, deep learning, and recent Transformer baselines across diverse asset classes?

\textbf{RQ2. Contribution of Adaptive Mechanisms.} How do specific architectural components, particularly the confidence-based meta-learner and the uncertainty quantification module, individually contribute to the framework's overall reliability and predictive power?

\textbf{RQ3. Efficacy of Dual-Channel Semantic Encoding.} Does the proposed dual-channel semantic encoding strategy yield significant performance gains compared to traditional single-channel or feature concatenation approaches in interpreting complex market context?

The following section details the ASTIF methodology designed to address these inquiries.

\section{Methodology}
\label{sec:methodology}
\begin{figure}[ht]
    \centering
    \includegraphics[width=1\linewidth]{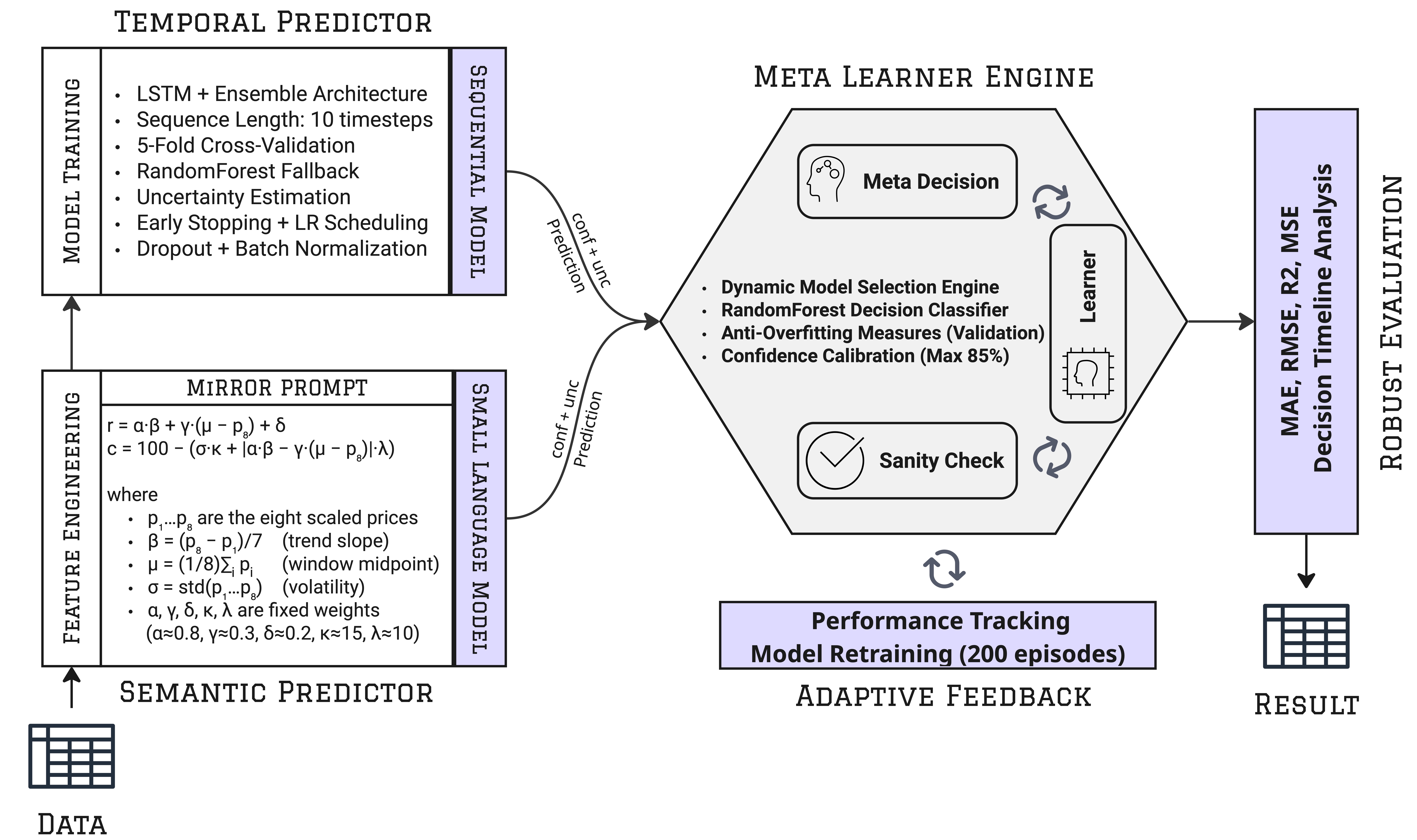}
    \caption{Structure of the ASTIF framework, illustrating the integration of MirrorPrompt SLM (dual-channel processing), LSTM-ML predictor, and meta-learner for dynamic cryptocurrency price forecasting.}
    \label{fig:astif_structure}
\end{figure}

We present ASTIF, a hierarchical ensemble architecture that synthesizes semantic language modeling, temporal neural networks, and meta-learning for reliable cryptocurrency price prediction. The framework addresses the fundamental challenge of market regime detection and model selection through three interconnected components: (i) a MirrorPrompt Small Language Model implementing dual-channel information processing, (ii) a hybrid LSTM-RF Temporal Learning Model with advanced feature engineering, and (iii) a confidence-calibrated meta-learner with anti-overfitting mechanisms. The ASTIF architecture demonstrated in (Figure~\ref{fig:astif_structure}) enables adaptive model selection based on market microstructure and uncertainty quantification, providing superior performance across volatile cryptocurrency markets.

\subsection{Problem Formulation and Notations}
We consider multi-asset time series forecasting under regime shifts and heteroscedastic noise. Let \(\{(\mathbf{X}_t, p_t)\}_{t=1}^{T}\) denote daily observations where \(\mathbf{X}_t \in [0,1]^d\) collects cross-asset signals (indices, ETFs, commodities, sentiment) and engineered technical indicators (momentum, SMA/EMA, volatility, RSI, Bollinger features) computed on Min-Max scaled series, and \(p_t \in [0,1]\) is the scaled closing price of the target asset. 

The objective is to produce calibrated next-step forecasts \(\hat y_{t+1} \in [0,1]\) that minimize predictive risk under nonstationary dynamics \(\mathcal{P}_t\):
\begin{equation}
\arg\min_{f}\; \mathbb{E}_{(\mathbf{X}_t, p_t) \sim \mathcal{P}_t}\, \ell\big(p_{t+1}, f(\mathbf{X}_{1:t}, p_{1:t})\big),
\end{equation}
where \(\ell\) denotes absolute or squared error loss.

We decompose forecasting into semantic and temporal experts. The semantic language model produces a return-confidence pair \((\hat r^{\mathrm{SLM}}_t, c_{\mathrm{SLM},t})\) from a dual-channel encoding of the last \(w\) prices and contextual features. The implied price prediction is:
\begin{equation}
\hat y^{\mathrm{SLM}}_{t+1} = p_t \left(1 + \frac{\hat r^{\mathrm{SLM}}_t}{100}\right),
\end{equation}
with uncertainty \(u_{\mathrm{SLM},t} = 1 - \min(c_{\mathrm{SLM},t},1)\).

The temporal learner produces predictions \(\hat y^{\mathrm{LSTM}}_{t+1}\) and \(\hat y^{\mathrm{RF}}_{t+1}\), combined as:
\begin{equation}
\hat y^{\mathrm{ML}}_{t+1} = \alpha \, \hat y^{\mathrm{LSTM}}_{t+1} + (1-\alpha)\, \hat y^{\mathrm{RF}}_{t+1},
\end{equation}
where \(\alpha \in (0,1)\). The disagreement-derived uncertainty is:
\begin{equation}
u_{\mathrm{ML},t} = \frac{\lvert \hat y^{\mathrm{LSTM}}_{t+1} - \hat y^{\mathrm{RF}}_{t+1} \rvert}{\max(\hat y^{\mathrm{LSTM}}_{t+1}, \hat y^{\mathrm{RF}}_{t+1})},
\end{equation}
yielding confidence \(c_{\mathrm{ML},t} = 1 - u_{\mathrm{ML},t}\).

A meta-selector \(\delta(\mathbf{z}_t) \in \{\mathrm{SLM}, \mathrm{ML}\}\) operates on a calibrated feature vector \(\mathbf{z}_t \in \mathbb{R}^{12}\) comprising confidences, uncertainties, inter-model disagreement, volatility proxies, trend strength, recent accuracies, and price momentum. The inter-model disagreement is computed as:
\begin{equation}
d_t = \frac{\lvert \hat y^{\mathrm{SLM}}_{t+1} - \hat y^{\mathrm{ML}}_{t+1} \rvert}{\max(\hat y^{\mathrm{SLM}}_{t+1}, \hat y^{\mathrm{ML}}_{t+1})}.
\end{equation}

The final prediction is:
\begin{equation}
\hat y_{t+1} = \hat y^{\delta(\mathbf{z}_t)}_{t+1},
\end{equation}
subject to feasibility \(\hat y_{t+1} \in [0,1]\) and a single-step change constraint \(\lvert \hat y_{t+1} - p_t \rvert / p_t \leq \tau_{\max}\). We seek to minimize expected loss:
\begin{equation}
\min \; \mathbb{E}\,\ell(p_{t+1}, \hat y_{t+1}),
\end{equation}
while ensuring probabilistic calibration. For confidence scores \(c\), the experimental coverage \(\hat p\) should satisfy \(|c - \hat p|\) small across bins, ensuring low calibration error. 
This formalization targets regime-adaptive, uncertainty-aware forecasts that remain valid in the bounded scaled domain and respect cross-asset dependencies:
\begin{equation}
\rho_{i,j}(t) = \mathrm{corr}(p_{i,t-w:t}, p_{j,t-w:t}).
\end{equation}

\subsection{Data Preprocessing and Feature Engineering}
The ASTIF framework employs a multi-source feature engineering pipeline operating on multi-asset time series data spanning January 2020 to December 2024. We collect seven diverse financial assets for systematic evaluation: five AI-focused cryptocurrency tokens and two traditional technology stocks. The AI-based cryptocurrency tokens include Fetch.AI (FET), SingularityNET (AGIX), Numeraire (NMR), Ocean Protocol (OCEAN), and Cortex (CTX), with historical daily closing prices retrieved from Coinmarketcap.com. The traditional technology stocks include NVIDIA (NVDA) and Microsoft (MSFT), serve as performance benchmarks, representing established technology companies with strong AI exposure. This diverse asset selection enables evaluation across different market segments and volatility patterns.

The dataset integrates over 29 cross-asset features across multiple categories to capture broad market dynamics. Cryptocurrency market data includes Bitcoin (BTC) and Ethereum (ETH) as foundational digital assets, alongside specialized token sectors: Decentralized Finance (DeFi) tokens, Non-Fungible Tokens (NFTs), Energy tokens, Gold tokens, Healthcare tokens, Asset Management tokens, and Lending \& Borrowing tokens, which exhibit documented spillover effects on AI token valuations. Traditional market indices include the S\&P Index, NASDAQ AI \& Robotics, and NASDAQ Internet Index as proxies for technological sector performance. Commodity markets include crude oil and gold prices as fundamental macroeconomic indicators, while sector-specific indices comprise Dow Jones Basic Materials (covering industrial metals, chemicals, and paper sectors) and Dow Jones Oil \& Gas (encompassing oil equipment, distribution, and alternative energy sectors). Thematic exchange-traded funds include iShares Global Clean Energy, iShares Robotics \& AI, ROBO ETFs, and Fintech Global ETFs, capturing innovation-driven market segments. The Blockchain Economic Index serves as a distributed ledger technology adoption indicator. Data sources include Coinmarketcap.com for cryptocurrency prices, the Energy Information Administration (EIA) for crude oil data, and the Thomson Reuters database for equity indices and ETF data.

For semantic feature engineering, we identify sentiment-driven and policy-related variables that capture market psychology and regulatory uncertainty. These semantic features include the Panic Index, Media Hype Index, Media Coverage Index, Fake News Index, and Infodemic Index, which quantify investor sentiment and information quality in digital asset markets. The GEPU index also serves as a macroeconomic state variable reflecting geopolitical and regulatory risks. These features are selected for semantic processing because they represent qualitative market conditions that language models can interpret contextually. Unlike purely numerical price movements, sentiment indicators encode narrative-driven market dynamics including fear and misinformation propagation. The semantic channel in our MirrorPrompt architecture transforms these features into natural language descriptions (e.g., "rising panic index amid regulatory concerns" or "declining media hype following market correction"), enabling the Small Language Model to employ its pre-trained understanding of financial narratives for prediction. All variables undergo unified Min-Max normalization to $[0,1]$ bounds via:
\begin{equation}
x'_i = \frac{x_i - \min(\mathbf{X})}{\max(\mathbf{X}) - \min(\mathbf{X})}
\end{equation}
ensuring numerical stability across heterogeneous feature scales. The technical indicator suite encompasses momentum-based features $M_{t,k} = (p_t - p_{t-k})/p_{t-k}$ for lags $k \in \{1,2,3,5,7\}$, trend-following indicators including Simple Moving Averages (SMA) and Exponentially Weighted Moving Averages (EMA) with varying window sizes. Volatility measures capture market microstructure through rolling standard deviation, Relative Strength Index (RSI), and Bollinger Bands, while cross-asset correlation features $\rho_{i,j}(t) = \text{corr}(p_{i,t-w:t}, p_{j,t-w:t})$ capture inter-market dependencies. The GEPU index \cite{caldara2022measuring} serves as a macroeconomic state variable with lagged transformations $\text{GEPU}_{t-k}$ for $k \in \{0,1,3,7\}$.

\subsection{The Semantic Predictor: MirrorPrompt Small Language Model}
\begin{figure}[ht]
    \centering
    \includegraphics[width=1\linewidth]{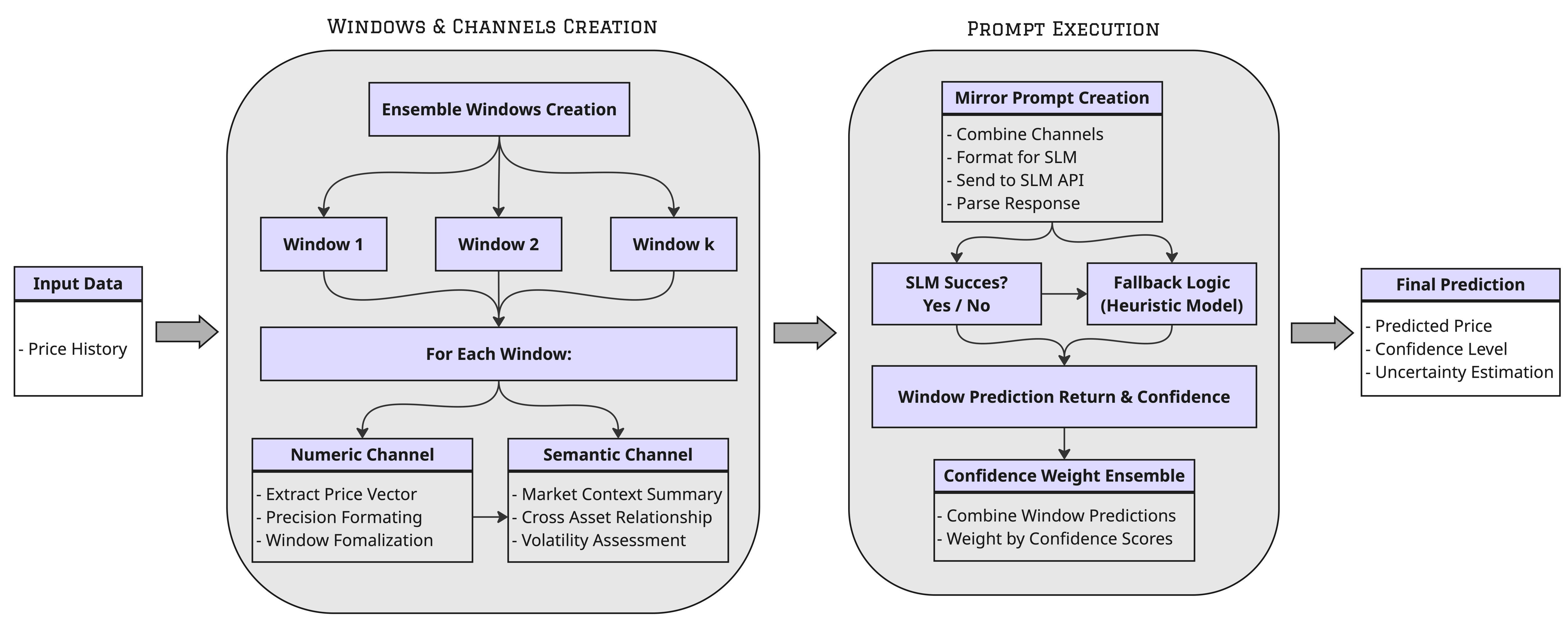}
    \caption{MirrorPrompt workflow structure, depicting ensemble window creation, dual-channel (numeric and semantic) processing, LLM querying, and confidence-weighted ensembling for cryptocurrency price forecasting.}
    \label{fig:mirrorprompt_structure}
\end{figure}

The MirrorPrompt architecture described in (Figure~\ref{fig:mirrorprompt_structure}) implements a novel dual-channel semantic learning paradigm that decomposes market analysis into complementary numeric and contextual processing streams. The numeric channel $\mathcal{N}(\mathbf{p}_{t-w:t})$ processes normalized price sequences through precision-encoded representations, while the semantic channel $\mathcal{S}(\mathbf{c}_t)$ generates contextual market narratives incorporating cross-asset dynamics, volatility regimes, and sentiment indicators. This decomposition enables the small language model to combine quantitative precision with qualitative market intuition.

The numeric channel encodes price sequences as high-precision decimal arrays $\mathbf{n}_t = [p_{t-w+1}, p_{t-w+2}, \ldots, p_t]$ where each element maintains 6-decimal precision to preserve MinMax-scaled microstructure information. The semantic channel constructs contextual narratives through template-driven text generation $\mathcal{S}(\mathbf{c}_t) = f_{\text{template}}(\Delta p_t, \sigma_t, \mathbf{I}_t, \mathbf{S}_t)$ incorporating percentage changes $\Delta p_t = (p_t - p_{t-1})/p_{t-1} \times 100$, volatility estimates $\sigma_t = \sqrt{\text{Var}(r_{t-w:t})}$, cross-asset indicators $\mathbf{I}_t$, and sentiment features $\mathbf{S}_t$.

The ensemble prediction mechanism operates over $K=3$ overlapping time windows $\{W_k\}_{k=1}^K$ with staggered endpoints to capture temporal dependencies at multiple scales. Each window generates a prediction-confidence pair $(\hat{r}_k, c_k) = \text{LLM}(\mathcal{N}(\mathbf{p}_{W_k}), \mathcal{S}(\mathbf{c}_{W_k}))$ through structured prompt engineering with deterministic sampling (temperature $\tau = 0$). The final ensemble prediction employs confidence-weighted averaging:
\begin{equation}
\hat{r}_{\text{SLM}} = \frac{\sum_{k=1}^{K} c_k \hat{r}_k}{\sum_{k=1}^{K} c_k}, \quad c_{\text{SLM}} = K^{-1}\sum_{k=1}^{K} c_k, \quad u_{\text{SLM}} = 1 - \min(c_{\text{SLM}}, 1)
\end{equation}
where uncertainty $u_{\text{SLM}}$ quantifies prediction reliability for meta-learning integration.

\subsection{The Temporal Predictor: Hybrid LSTM-RF Architecture}
The temporal learning component implements a hierarchical ensemble architecture combining sequence-aware LSTM networks with feature-interaction-sensitive Random Forest models. This hybrid approach addresses the dual challenges of long-term temporal dependency modeling and complex cross-feature interaction learning inherent in multi-asset cryptocurrency prediction tasks. The architecture employs full-dataset training paradigms with advanced regularization techniques to maximize information utilization while preventing overfitting.

The LSTM architecture implements a three-layer hierarchical structure $(64 \to 32 \to 16)$ with batch normalization and dropout regularization. Each layer operates on sequences $\mathbf{X}_t \in \mathbb{R}^{L \times d}$ where $L=10$ represents the sequence length and $d$ denotes feature dimensionality. The LSTM cell dynamics follow:
\begin{equation}
\begin{aligned}
f_t &= \sigma(W_f \cdot [h_{t-1}, x_t] + b_f), \quad i_t = \sigma(W_i \cdot [h_{t-1}, x_t] + b_i) \\
\tilde{C}_t &= \tanh(W_C \cdot [h_{t-1}, x_t] + b_C), \quad C_t = f_t \odot C_{t-1} + i_t \odot \tilde{C}_t \\
o_t &= \sigma(W_o \cdot [h_{t-1}, x_t] + b_o), \quad h_t = o_t \odot \tanh(C_t)
\end{aligned}
\end{equation}
where $\sigma$ denotes the sigmoid function and $\odot$ represents element-wise multiplication.

The Random Forest component employs 100 estimators with controlled depth and minimum sample constraints to prevent overfitting. Ensemble uncertainty quantification uses prediction disagreement: $u_{\text{ML}} = |\hat{y}_{\text{LSTM}} - \hat{y}_{\text{RF}}| / \max(\hat{y}_{\text{LSTM}}, \hat{y}_{\text{RF}})$, with confidence derived as $c_{\text{ML}} = 1 - u_{\text{ML}}$. The final ML prediction uses confidence-weighted averaging: $\hat{y}_{\text{ML}} = (0.7 \hat{y}_{\text{LSTM}} + 0.3 \hat{y}_{\text{RF}})$ based on experimental performance validation.

\subsection{Confidence-Calibrated Meta-Learning}
The meta-learning component implements a RandomForest classifier with conservative regularization parameters (20 estimators, max depth 5) trained on a 12-dimensional feature vector $\mathbf{z}_t \in \mathbb{R}^{12}$ encoding model predictions, confidence scores, uncertainty estimates, and market context. The feature extraction function $\phi(\hat{y}_{\text{SLM}}, c_{\text{SLM}}, u_{\text{SLM}}, \hat{y}_{\text{ML}}, c_{\text{ML}}, u_{\text{ML}}, \mathbf{m}_t)$ incorporates:

$\mathbf{z}_t = [c_{\text{SLM}}, c_{\text{ML}}, u_{\text{SLM}}, u_{\text{ML}}, |\hat{y}_{\text{SLM}} - \hat{y}_{\text{ML}}|/\max(\hat{y}_{\text{SLM}}, \hat{y}_{\text{ML}}), |c_{\text{SLM}} - c_{\text{ML}}|, |u_{\text{SLM}} - u_{\text{ML}}|, \sigma_t, \rho_{\text{trend}}, \alpha_{\text{SLM}}, \alpha_{\text{ML}}, \Delta p_t]^\top$
where $\sigma_t$ represents market volatility, $\rho_{\text{trend}}$ captures trend strength, $\alpha_{\text{SLM/ML}}$ denote recent model accuracy, and $\Delta p_t$ represents price momentum.

Anti-overfitting mechanisms include temporal validation splits with 30\% holdout ratio, overfit detection through train-validation accuracy gap monitoring ($\Delta_{acc} > 0.15$ threshold), and conservative confidence calibration. The calibration function transforms raw meta-probabilities $p_{\text{raw}}$ through regime-dependent scaling:
\begin{equation}
p_{\text{cal}} = \begin{cases}
0.4 + 0.2(p_{\text{raw}} - 0.5) & \text{if } \bar{c} < 0.6 \\
0.3 + 0.4p_{\text{raw}} & \text{if } 0.6 \leq \bar{c} < 0.8 \\
0.2 + 0.6p_{\text{raw}} & \text{if } \bar{c} \geq 0.8
\end{cases}
\end{equation}
where $\bar{c} = (c_{\text{SLM}} + c_{\text{ML}})/2$ represents average model confidence, and calibrated confidence is capped at $p_{\text{cal}} \leq 0.85$ to prevent overconfidence.

\subsection{Adaptive Integration with Uncertainty-Aware Decision Making}
The adaptive integration algorithm synthesizes predictions through a hierarchical decision process incorporating meta-learned model selection, confidence calibration, and sanity checking mechanisms. The decision function $\delta: \mathbb{R}^{12} \to \{\text{SLM}, \text{ML}\}$ operates on the meta-feature vector $\mathbf{z}_t$ to select the optimal predictor for current market conditions.

The selection mechanism employs both learned and rule-based components. When sufficient training data exists ($|\mathcal{E}| \geq 50$ episodes), the RandomForest meta-classifier produces selection probabilities $P(\text{SLM}|\mathbf{z}_t)$ and $P(\text{ML}|\mathbf{z}_t) = 1 - P(\text{SLM}|\mathbf{z}_t)$. Otherwise, rule-based selection applies high-confidence thresholds ($c_{\text{SLM}} \geq 0.9 \land u_{\text{SLM}} < 0.2$) or uncertainty-based tiebreaking for similar confidence scenarios.

Sanity checking validates predictions within bounded feasibility regions. For MinMax-scaled data, predictions must satisfy $\hat{y}_i \in [0,1]$ and relative change constraints $|\hat{y}_i - p_t|/p_t \leq \tau_{\text{max}}$ where $\tau_{\text{max}} = 0.5$ represents the maximum acceptable single-step change threshold. Violations trigger model switching or conservative prediction capping.

The final prediction integrates model outputs through confidence-weighted selection:
\begin{equation}
\hat{y}_{\text{final}} = \begin{cases}
\hat{y}_{\text{SLM}} & \text{if } \delta(\mathbf{z}_t) = \text{SLM} \land \mathcal{V}(\hat{y}_{\text{SLM}}) = \text{True} \\
\hat{y}_{\text{ML}} & \text{if } \delta(\mathbf{z}_t) = \text{ML} \land \mathcal{V}(\hat{y}_{\text{ML}}) = \text{True} \\
\arg\min_{\hat{y} \in \{\hat{y}_{\text{SLM}}, \hat{y}_{\text{ML}}\}} |\hat{y} - p_t|/p_t & \text{otherwise (sanity override)}
\end{cases}
\end{equation}
where $\mathcal{V}(\cdot)$ denotes the sanity validation function and the fallback selects the prediction with smaller relative deviation from current price.

\subsection{Training Objective and Optimization}
The ASTIF framework employs component-specific training objectives optimized through distinct procedures for temporal learning and meta-learning subsystems. The semantic learning model operates through deterministic prompt-based inference without gradient-based training, relying instead on structured prompt engineering to elicit predictions from the pre-trained language model.

The temporal learning component optimizes prediction accuracy through regression loss minimization. The LSTM network trains on mean squared error loss:
\begin{equation}
\mathcal{L}_{\text{LSTM}} = \frac{1}{N}\sum_{i=1}^{N}(y_i - \hat{y}_{\text{LSTM},i})^2
\end{equation}
where $N$ represents training samples, $y_i$ denotes target prices, and $\hat{y}_{\text{LSTM},i}$ are LSTM predictions. The optimization employs Adam optimizer with learning rate $\alpha = 0.001$, batch size 32, and early stopping monitoring validation loss with patience of 15 epochs for cross-validation and 25 epochs for final training. Dropout regularization (rate 0.2) and batch normalization prevent overfitting in the three-layer hierarchical architecture. The Random Forest component trains through entropy-based splitting criteria without explicit loss function optimization, employing bootstrap aggregation with 100 trees, maximum depth of 20, and minimum sample constraints to prevent overfitting.

The meta-learning classifier optimizes decision accuracy through cross-entropy loss:
\begin{equation}
\mathcal{L}_{\text{meta}} = -\frac{1}{T}\sum_{t=1}^{T}\left[y_t^{\text{SLM}}\log P(\text{SLM}|\mathbf{z}_t) + y_t^{\text{ML}}\log P(\text{ML}|\mathbf{z}_t)\right]
\end{equation}
where $y_t^{\text{SLM}} \in \{0,1\}$ indicates whether SLM achieves lower error at timestep $t$, $y_t^{\text{ML}} = 1 - y_t^{\text{SLM}}$, and $P(\text{SLM}|\mathbf{z}_t)$ represents the meta-classifier's predicted selection probability. Training labels derive from retrospective evaluation comparing $\text{MAE}_{\text{SLM}}(t)$ and $\text{MAE}_{\text{ML}}(t)$ on validation data, assigning the superior model as ground truth.

The unified MinMax scaling architecture ensures numerical stability across heterogeneous feature scales, mapping all inputs to $[0,1]$ bounds through $x'_i = (x_i - \min(\mathbf{X})) / (\max(\mathbf{X}) - \min(\mathbf{X}))$. This standardization prevents gradient instability in LSTM training and enables consistent cross-asset model deployment without architecture modifications. Temporal data splitting maintains chronological integrity with 80\% allocated to training and 20\% to testing, where the last 10\% of the training portion serves as an internal validation slice for early stopping and hyperparameter calibration.

\section{Experimental Setup}
\label{sec:experiments}
This section details the protocol used to evaluate the ASTIF described in Section~\ref{sec:methodology}. We align the experimental design with the architecture’s three components: MirrorPrompt SLM, LSTM–RF temporal learner, and the confidence–calibrated meta-learner, so that model behavior under different market regimes can be assessed consistently and without leakage.

\subsection{Dataset Description}
We evaluate on daily multi-asset time series spanning January 2020 to December 2024. The target set includes major and AI-linked cryptocurrencies (e.g., FET, Ocean, NMR, AGIX ), alongside a superset of cross-asset variables that encode equity indices, sectoral and thematic ETFs (clean energy, robotics and AI), commodities (gold, crude oil), and sentiment sources (panic and media coverage indices). The GEPU index \citep{caldara2022measuring} is merged on dates to capture macroeconomic conditions. All variables are transformed with a unified Min–Max normalization in the \([0,1]\) range, ensuring numerically stable training on heterogeneous scales as specified in Section~\ref{sec:methodology}. The technical indicators used by the temporal learner, such as momentum, moving averages (SMA/EMA), rolling volatility, RSI, and Bollinger Bands, are computed in scaled series within the learning pipeline to avoid data duplication and preserve chronological integrity.

\subsection{Baseline Models}
We compare ASTIF against strong non-sequential baselines trained on the same features and evaluated on the identical forecast horizon: Machine Learning, Deep Learning and Transformer-based models with standardized features. These baselines provide a spectrum of capacity for non-linear function approximation and feature interactions without explicit temporal state. Where relevant, we align the evaluation window to match the hybrid system’s effective prediction coverage, ensuring that all models are scored on the same target samples.

\subsection{Evaluation Metrics}
Primary accuracy is quantified using mean absolute error and root mean square error, and we report the coefficient of determination for explanatory power. For a sequence of true values \(\{y_i\}_{i=1}^n\) and predictions \(\{\hat{y}_i\}_{i=1}^n\), we compute
\[
	{MAE} = \frac{1}{n}\sum_{i=1}^{n} |y_i - \hat{y}_i|,\quad
	{RMSE} = \sqrt{\frac{1}{n}\sum_{i=1}^{n} (y_i - \hat{y}_i)^2},\quad
	{R}^2 = 1 - \frac{\sum_{i=1}^{n} (y_i - \hat{y}_i)^2}{\sum_{i=1}^{n} (y_i - \bar{y})^2}.
\]
Following Section~\ref{sec:methodology}, we assess uncertainty calibration through reliability diagrams that compare predicted confidence with experimental coverage and summarize deviation by a calibration error \(\text{CE} = \sum_{j=1}^{B} |p_j - \hat{p}_j|\, n_j/n\), where \(p_j\) and \(\hat{p}_j\) denote predicted and observed accuracies within bin \(j\). We further monitor meta-decision quality via temporal cross-validation, recording the fraction of steps on which the selected model attains lower error than the alternative.

\subsection{Implementation Details and Hyperparameters}
All models operate exclusively in the scaled $[0,1]$ space to ensure numerical stability and cross-asset comparability. The dataset employs a chronological split with 80\% allocated to training and 20\% to testing. Within the training portion, the last 10\% serves as an internal validation slice for early stopping and calibration in the temporal learner, as well as episodic validation in the meta-learner. These validation protocols mirror the anti-overfitting safeguards described in Section~\ref{sec:methodology}. Some analyses also employ a forward-prediction protocol where a small initial context enables sequential forecasting across the remainder of the series to characterize end-to-end coverage.

The SLM component employs the Gemma-3-1b-it model deployed via LM Studio API with a 30-second timeout for response generation. The model operates with max\_tokens=150 (extended to 600 for reasoning models) and implements the MirrorPrompt strategy. MirrorPrompt operates with dual channels: a numeric channel comprising the last $w=8$ scaled prices (six-decimal precision) and a semantic channel that summarizes short-horizon momentum, volatility, cross-asset signals, and sentiment indicators. Deterministic sampling is enforced (temperature $\tau=0$) to ensure reproducibility. An overlapping ensemble of $K=3$ windows provides confidence-weighted aggregation of return predictions as defined in Section~\ref{sec:methodology}. Predicted returns are mapped to next-step scaled prices through multiplicative updates relative to the current price level.

The temporal learner combines a sequence model and a tree ensemble to capture both temporal dependencies and feature interactions. The LSTM architecture uses a three-layer hierarchy with 64, 32, and 16 units respectively, incorporating batch normalization and dropout of 0.2 for regularization. Training employs the Adam optimizer with a learning rate of 0.001 on sequences of length $L=10$. The complementary Random Forest provides interaction-sensitive structural modeling with 100 trees, depth control, and bootstrap sampling. Final ML predictions employ an experimentally validated convex combination of the two branches, with weights favoring the sequence model (0.7 LSTM, 0.3 RF).

The meta-learner employs a Random Forest classifier with conservative regularization parameters: 20 estimators, maximum depth of 5, and minimum sample constraints to prevent overfitting. The classifier operates on a 12-dimensional feature vector comprising model confidences, uncertainties, relative disagreement, and market context features. Meta-probabilities undergo conservative calibration using the regime-dependent mapping described in Section~\ref{sec:methodology}. Decision confidence is explicitly capped to prevent overconfidence in volatile market conditions.

To avoid pathological behavior in the scaled domain, all forecasts undergo feasibility validation. Predictions must remain within $[0,1]$ bounds, and single-step relative changes cannot exceed a threshold of 0.5. When a candidate prediction violates these feasibility constraints, the system either switches to the alternative model or applies conservative forecast capping, as detailed in the adaptive integration rules of Section~\ref{sec:methodology}.

All experiments employ unified Min-Max scaling and chronological data splits to ensure temporal consistency. The SLM uses deterministic decoding to eliminate sampling variability. The temporal learner and baseline models use fixed random seeds for reproducibility. The meta-learner employs temporal holdout validation with conservative calibration to ensure stable performance. Unless otherwise noted, all results are reported in the scaled $[0,1]$ space to maintain comparability across assets and feature regimes.

\section{Results}
\label{sec:results}
This section presents experimental findings addressing three research questions: comparative performance evaluation against established machine learning, deep learning, and transformer baselines, quantitative assessment of architectural component contributions through systematic ablation, and analysis of semantic input configuration effects on forecasting performance. Results are derived from experiments conducted across seven diverse financial assets, including AI-focused cryptocurrencies (FET, AGIX, NMR, OCEAN, CTX) and traditional equities (MSFT, NVDA).

\subsection{Comparative Performance against Baselines}
\label{sec:rq1_results}
The first research question investigates how the ASTIF framework performs compare to traditional machine learning models including Decision Tree (DTC), Support Vector Machine (SVM), Random Forest (RFC), and AdaBoost (ADB) as well as deep learning architectures such as LSTM and CNN-LSTM, and transformer-based models including the standard Transformer (TF), Informer, and Temporal Fusion Transformer (TFT). The comparison focuses on forecasting accuracy and prediction stability across a diverse set of asset classes.

The results clearly reveal that ASTIF consistently outperforms all competing models, achieving the lowest error rates across the portfolio. specifically, MAE values ranging from 0.0073 for traditional equities to 0.0133 for more volatile cryptocurrency assets, as summarized in (Table~\ref{tab:model_comparison}). The traditional machine learning approaches, SVM exhibits the poorest performance and higher instability, with MAE values between 0.0595 to 0.1060, almost five times higher than ASTIF. Deep learning approaches deliver moderate performance, while transformer-based architecture are the strongest among the baselines, with Informer performing best for most asset classes.

\begin{table}[htbp]
\centering
\tiny
\setlength{\tabcolsep}{1pt}
\caption{Performance comparison across asset classes. Best values are in \textbf{bold}.}
\label{tab:model_comparison}
\resizebox{\textwidth}{!}{%
\begin{tabular*}{\textwidth}{@{\extracolsep{\fill}} l *{14}{S[table-format=0.4]}}
\toprule
\textbf{Model} & \multicolumn{2}{c}{\textbf{CTX}} & \multicolumn{2}{c}{\textbf{FET}} & \multicolumn{2}{c}{\textbf{NMR}} & \multicolumn{2}{c}{\textbf{OCEAN}} & \multicolumn{2}{c}{\textbf{AGIX}} & \multicolumn{2}{c}{\textbf{MSFT}} & \multicolumn{2}{c}{\textbf{NVDA}} \\
\cmidrule(lr){2-3} \cmidrule(lr){4-5} \cmidrule(lr){6-7} \cmidrule(lr){8-9} \cmidrule(lr){10-11} \cmidrule(lr){12-13} \cmidrule(lr){14-15}
& \textbf{MAE} & \textbf{RMSE} & \textbf{MAE} & \textbf{RMSE} & \textbf{MAE} & \textbf{RMSE} & \textbf{MAE} & \textbf{RMSE} & \textbf{MAE} & \textbf{RMSE} & \textbf{MAE} & \textbf{RMSE} & \textbf{MAE} & \textbf{RMSE} \\
\midrule
\multicolumn{15}{l}{\textit{Machine Learning Models}} \\
\quad DTC & 0.0258 & 0.0642 & 0.0246 & 0.0976 & 0.0225 & 0.0415 & 0.0206 & 0.0461 & 0.0269 & 0.0854 & 0.0368 & 0.0934 & 0.0476 & 0.1267 \\
\quad RFC & 0.0347 & 0.0859 & 0.0269 & 0.1028 & 0.0231 & 0.0513 & 0.0351 & 0.0822 & 0.0239 & 0.0862 & 0.0376 & 0.0923 & 0.0511 & 0.1371 \\
\quad ADB & 0.0539 & 0.0924 & 0.0349 & 0.1054 & 0.0443 & 0.0552 & 0.0370 & 0.0490 & 0.0401 & 0.0892 & 0.0539 & 0.0972 & 0.0630 & 0.1412 \\
\quad SVM & 0.0818 & 0.1151 & 0.0770 & 0.1218 & 0.0605 & 0.0739 & 0.0595 & 0.0718 & 0.0826 & 0.1236 & 0.0845 & 0.1308 & 0.1060 & 0.1601 \\
\midrule
\multicolumn{15}{l}{\textit{Deep Learning Models}} \\
\quad LSTM & 0.0247 & \textbf{0.0422} & 0.0343 & 0.0464 & 0.0246 & 0.0427 & 0.0192 & 0.0264 & 0.0253 & 0.0388 & 0.0243 & 0.0462 & 0.0142 & 0.0189 \\
CNN-LSTM & 0.0317 & 0.0467 & 0.0364 & 0.0485 & 0.0282 & 0.0445 & 0.0206 & 0.0279 & 0.0276 & 0.0407 & 0.0274 & 0.0517 & 0.0149 & 0.0196 \\
\midrule
\multicolumn{15}{l}{\textit{Transformer Models}} \\
\quad TF & 0.0554 & 0.0685 & 0.0449 & 0.0561 & 0.0301 & 0.0467 & 0.0432 & 0.0512 & 0.0313 & 0.0451 & 0.0258 & 0.0465 & 0.0204 & 0.0256 \\
\quad Informer & 0.0251 & 0.0423 & 0.0324 & \textbf{0.0447} & 0.0245 & 0.0426 & 0.0181 & \textbf{0.0256} & 0.0236 & \textbf{0.0378} & 0.0299 & 0.0529 & 0.0138 & 0.0182 \\
\quad TFT & 0.0248 & 0.0424 & 0.0325 & \textbf{0.0447} & 0.0246 & 0.0427 & 0.0191 & 0.0264 & 0.0236 & 0.0380 & 0.0283 & 0.0508 & 0.0138 & 0.0184 \\
\midrule
\textbf{ASTIF} & \textbf{0.0226} & 0.0576 & \textbf{0.0133} & 0.0521 & \textbf{0.0158} & \textbf{0.0305} & \textbf{0.0168} & 0.0421 & \textbf{0.0213} & 0.0558 & \textbf{0.0222} & \textbf{0.0440} & \textbf{0.0073} & \textbf{0.0130} \\
\bottomrule
\end{tabular*}%
}
\end{table}

\begin{figure}[ht]
    \centering
    \includegraphics[width=1\columnwidth]{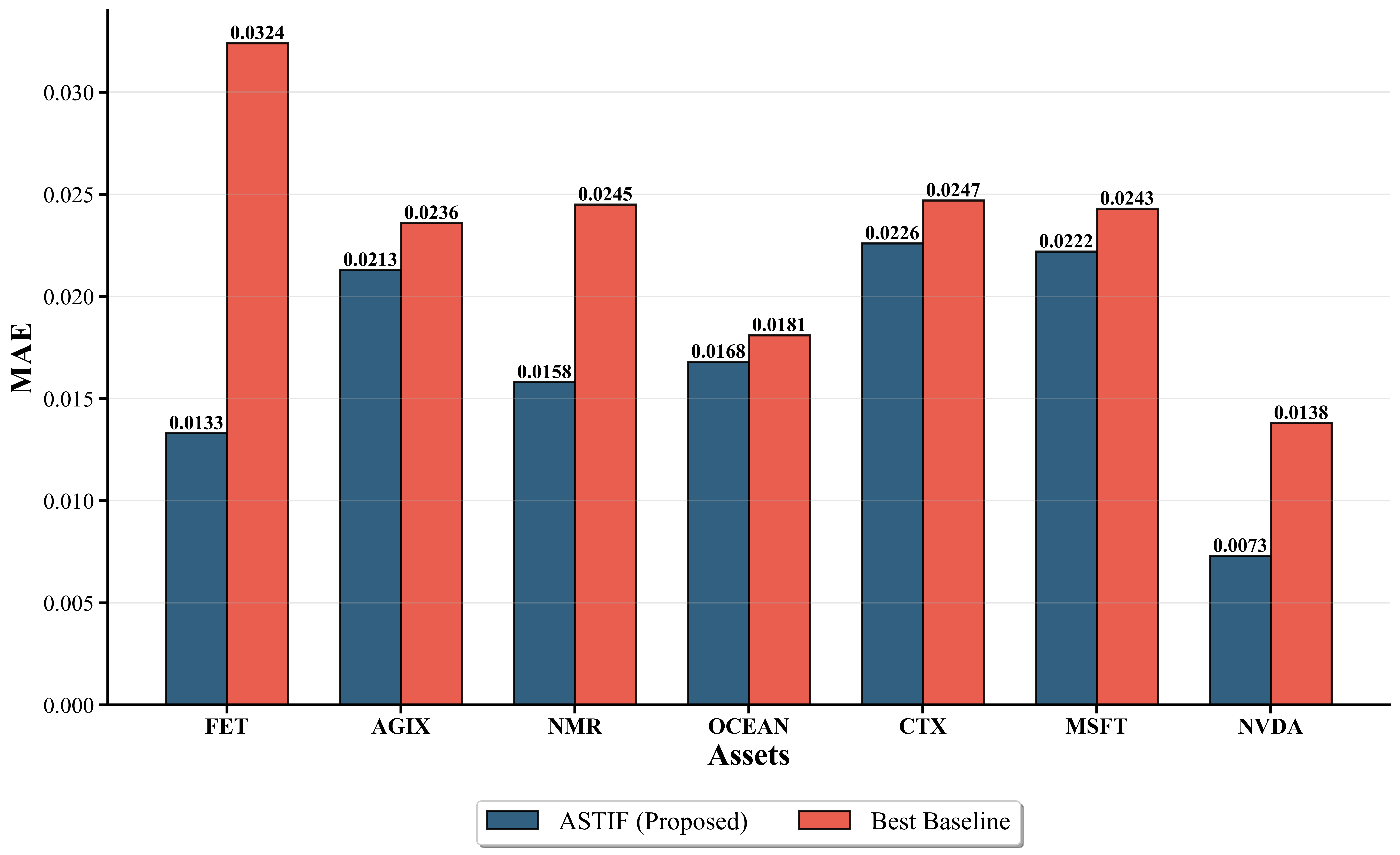}
    \caption{Forecasting improvement margins over optimal baseline models.}
\label{fig:astif_comparison}
\end{figure}

The magnitude of improvement over the most competitive baseline differs noticeably across asset, ranging from 7.2\% to 59.0\% as summarized in (Table~\ref{tab:astif_comparison}) and illustrated in (Figure~\ref{fig:astif_comparison}). Assets in the cryptocurrency category benefit the most, achieving substantial error reductions of 59.0\% and 35.5\% in representative examples, which underscores the ASTIF’s advantage in high volatile and complex market conditions. Traditional equity assets, in contrast, experience smaller but steady improvements, generally between 8.5\% and 8.6\%, reflecting the more predictable behavior of these markets and the relatively strong performance of existing models.

\begin{figure}[ht]
    \centering
    \includegraphics[width=1\columnwidth]{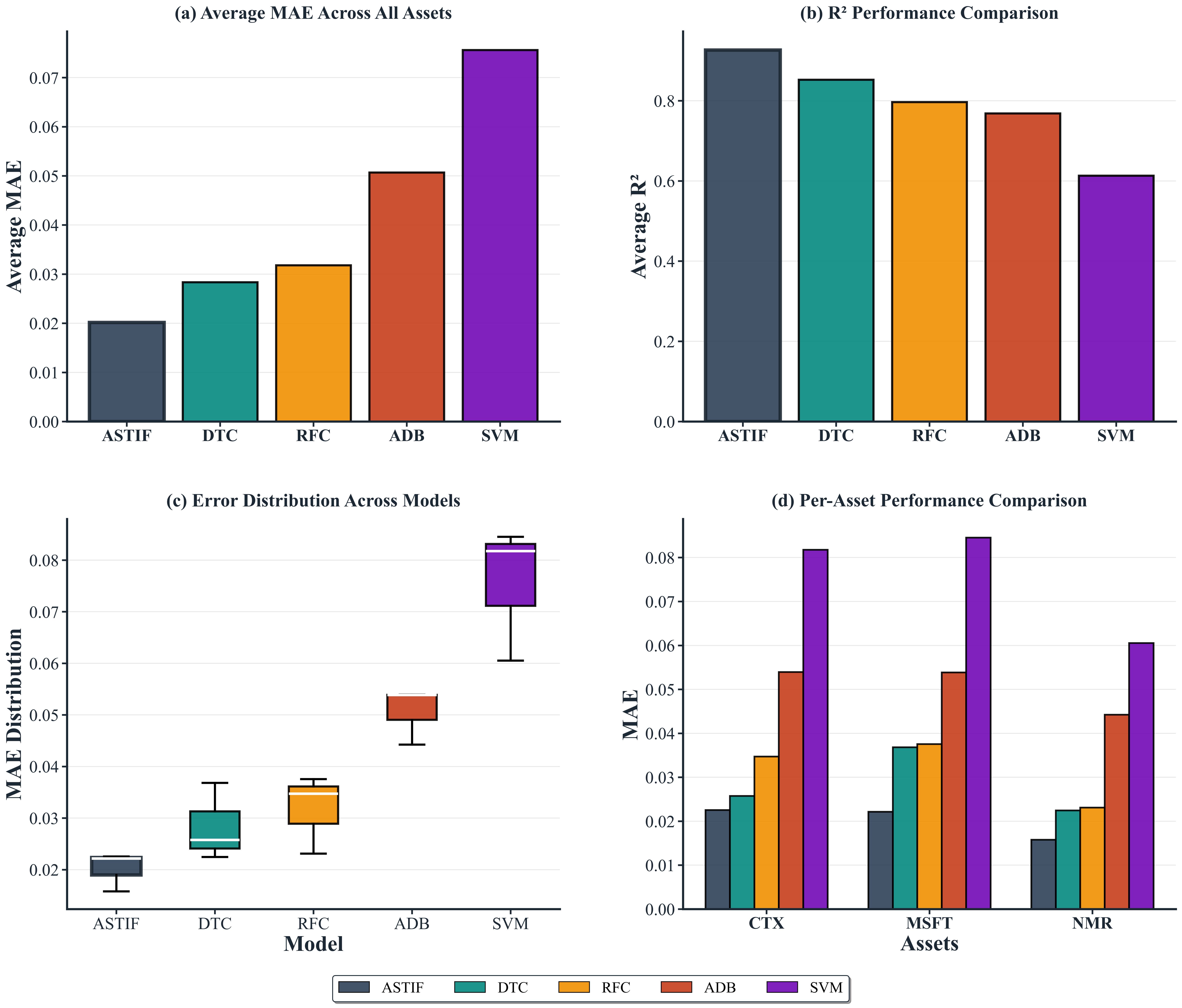}
    \caption{Comparative error analysis and stability metrics across model architectures.}
    \label{fig:model_performance}
\end{figure}

Multi-metric analysis reveals that ASTIF achieves the lowest average MAE among all machine learning architectures, maintains superior R² coefficients indicating better predictive accuracy, exhibits consistently tight error distribution with minimal variance, and confirms systematic outperformance across representative assets, as shown in the four-panel comparison of (Figure~\ref{fig:model_performance}). 

\begin{table}[htbp]
\centering
\caption{Relative MAE reduction against best performing baselines. Best values are in \textbf{bold}.}
\label{tab:astif_comparison}
\begin{tabular}{lcccc}
\toprule
\textbf{Asset} & \textbf{Best Baseline} & \textbf{Baseline MAE} & \textbf{ASTIF MAE} & \textbf{Imp - \%} \\
\midrule
FET & Informer & 0.0324 & \textbf{0.0133} & 59.0 \\
NVDA & Informer & 0.0138 & \textbf{0.0073} & 47.1 \\
NMR & Informer & 0.0245 & \textbf{0.0158} & 35.5 \\
AGIX & Informer & 0.0236 & \textbf{0.0213} & 9.7 \\
MSFT & LSTM & 0.0243 & \textbf{0.0222} & 8.6 \\
CTX & LSTM & 0.0247 & \textbf{0.0226} & 8.5 \\
OCEAN & Informer & 0.0181 & \textbf{0.0168} & 7.2 \\
\bottomrule
\end{tabular}
\end{table}

\begin{figure}[ht]
    \centering
    \includegraphics[width=1\columnwidth]{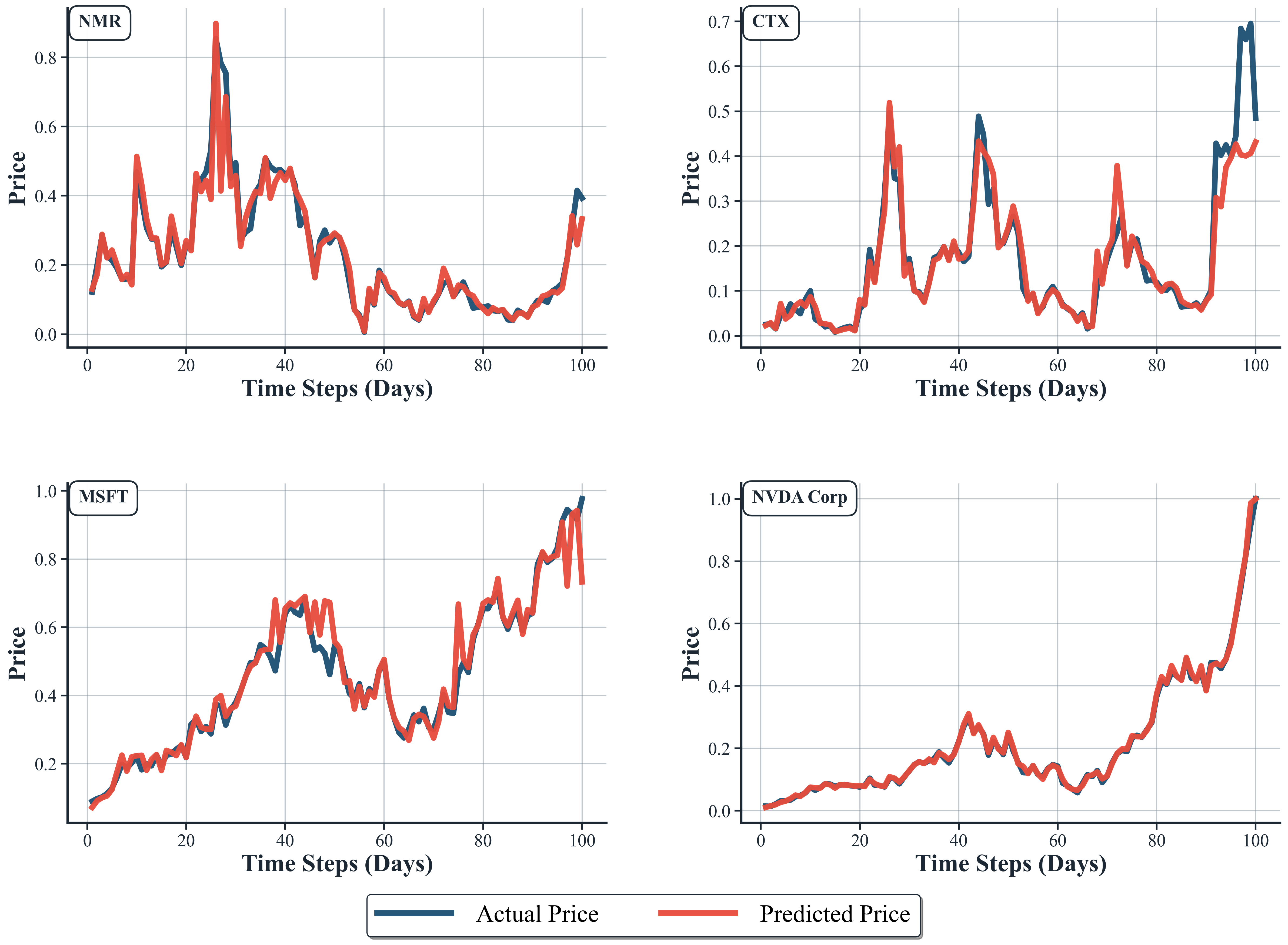}
    \caption{Actual versus predicted price trajectories for representative assets.}
    \label{fig:actual_vs_pred}
\end{figure}

\begin{figure}[ht]
    \centering
    \includegraphics[width=1\columnwidth]{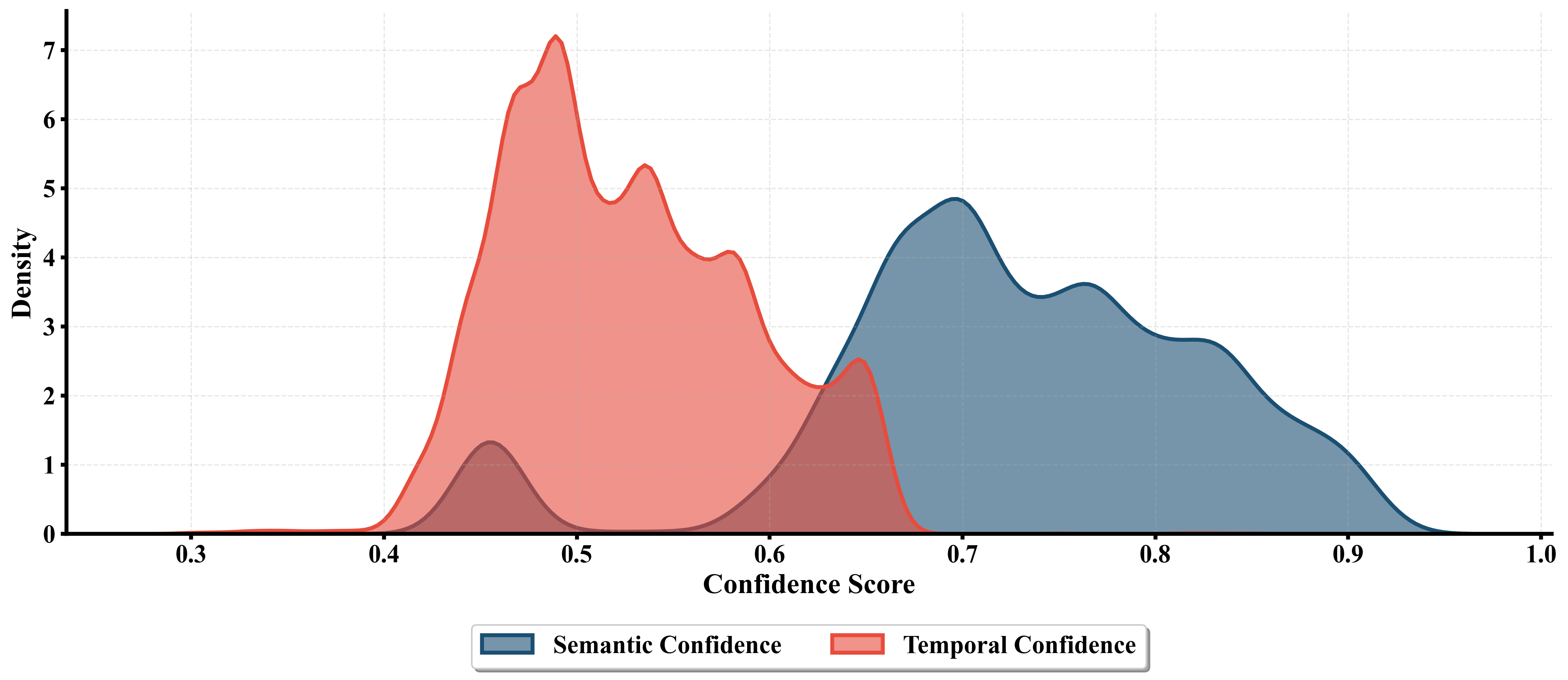}
    \caption{Meta-learner confidence distribution between semantic and temporal channels.}
    \label{fig:meta_confidence}
\end{figure}

Predicted values exhibit close alignment with actual prices across representative assets, as demonstrated in (Figure~\ref{fig:actual_vs_pred}). The meta-learner generates higher prediction confidence when semantic and temporal components align, indicating effective integration of heterogeneous information sources and providing interpretable reliability estimates, as shown in (Figure~\ref{fig:meta_confidence}). Taken together, these results indicate that ASTIF consistently enhances predictive accuracy across all examined assets, with its relative advantage becoming increasingly pronounced as market uncertainty and complexity grow.

\subsection{Impact of Adaptive Components (Ablation Study)}
\label{sec:rq2_results}
The second research question quantifies the contribution of individual components within the ASTIF architecture using a systematic ablation design. The SLM-based semantic predictor, LSTM-based temporal predictor, adaptive meta-learner selection mechanism, and uncertainty quantification framework are removed individually, and changes in predictive accuracy and reliability are measured.

\begin{figure}[ht]
    \centering
    \includegraphics[width=1\columnwidth]{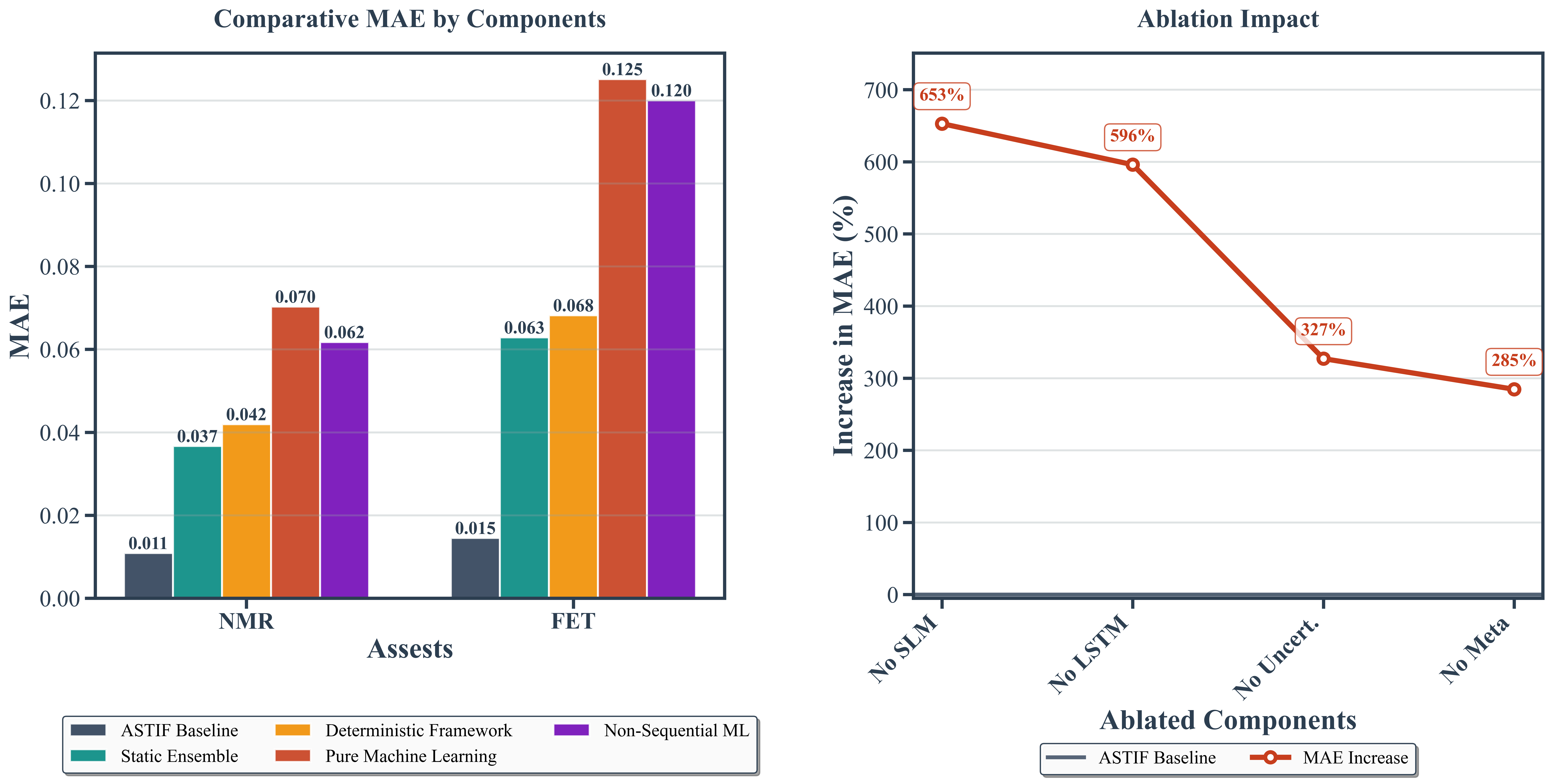}
    \caption{Performance degradation magnitude across ablation scenarios.}
    \label{fig:component_ablation}
\end{figure}

The results of this analysis are reported in (Table~\ref{tab:ablation_compact}). Removing the SLM semantic channel leads to the largest increase in MAE, from 0.0109 to 0.0703 for one cryptocurrency and from 0.0146 to 0.1252 for the other. Ablation of the LSTM component produces a comparable pattern, with MAE values rising to 0.0618 and 0.1201. Excluding the uncertainty quantification component results in MAE increases to 0.0420 and 0.0682, while removing the adaptive meta-learner selection mechanism yields smaller, though still observable, increases to 0.0367 and 0.0629. These findings are further visualized in (Figure~\ref{fig:component_ablation} Comparative MAE by components).

\begin{table}[htbp]
\centering
\caption{Architectural contribution analysis via component ablation. Best values are in \textbf{bold}.}

\label{tab:ablation_compact}
\small
\begin{tabular}{llcccc}
\toprule
\textbf{Configuration} & \textbf{Removed} & \multicolumn{2}{c}{\textbf{NMR}} & \multicolumn{2}{c}{\textbf{FET}} \\
\cmidrule(lr){3-4} \cmidrule(lr){5-6}
& & \textbf{MAE} & \textbf{RMSE} & \textbf{MAE} & \textbf{RMSE} \\
\midrule
ASTIF Baseline & None & \textbf{0.0109} & \textbf{0.0180} & \textbf{0.0146} & \textbf{0.0298} \\
Pure Machine Learning & SLM Channel & 0.0703 & 0.0881 & 0.1252 & 0.2459 \\
Non-Sequential ML & LSTM Modeling & 0.0618 & 0.0791 & 0.1201 & 0.2308 \\
Deterministic Frame. & Uncertainty & 0.0420 & 0.0649 & 0.0682 & 0.1726 \\
Static Ensemble & Meta-Learner & 0.0367 & 0.0462 & 0.0629 & 0.1240 \\
\bottomrule
\end{tabular}
\end{table}

Component ablation analysis reveals the critical interdependence of ASTIF's architectural elements. Removing the SLM semantic channel produces the most severe degradation, with MAE increasing by 653\% on average across assets. LSTM removal follows closely at 596\%, confirming the importance of temporal modeling (Figure~\ref{fig:component_ablation} Ablation impact). The uncertainty quantification mechanism proves similarly essential, its absence leads to a 327\% rise in prediction error. Even the adaptive meta-learner, while showing the smallest impact, still causes a substantial 285\% error increase when removed. These results indicate that ASTIF's superior performance stems from synergistic integration of all architectural components, with semantic processing and temporal modeling emerging as the most indispensable elements.

\subsection{Evaluation of Semantic Encoding Strategies}
\label{sec:rq3_results}
The third research question examines how variations in semantic input formulation affect ASTIF's forecasting capability through analysis of dual-channel versus single-channel semantic encoding, market context integration methods, and sensitivity analysis with temporal window and ensemble size configurations.

\begin{figure}[ht]
    \centering
    \includegraphics[width=1\linewidth]{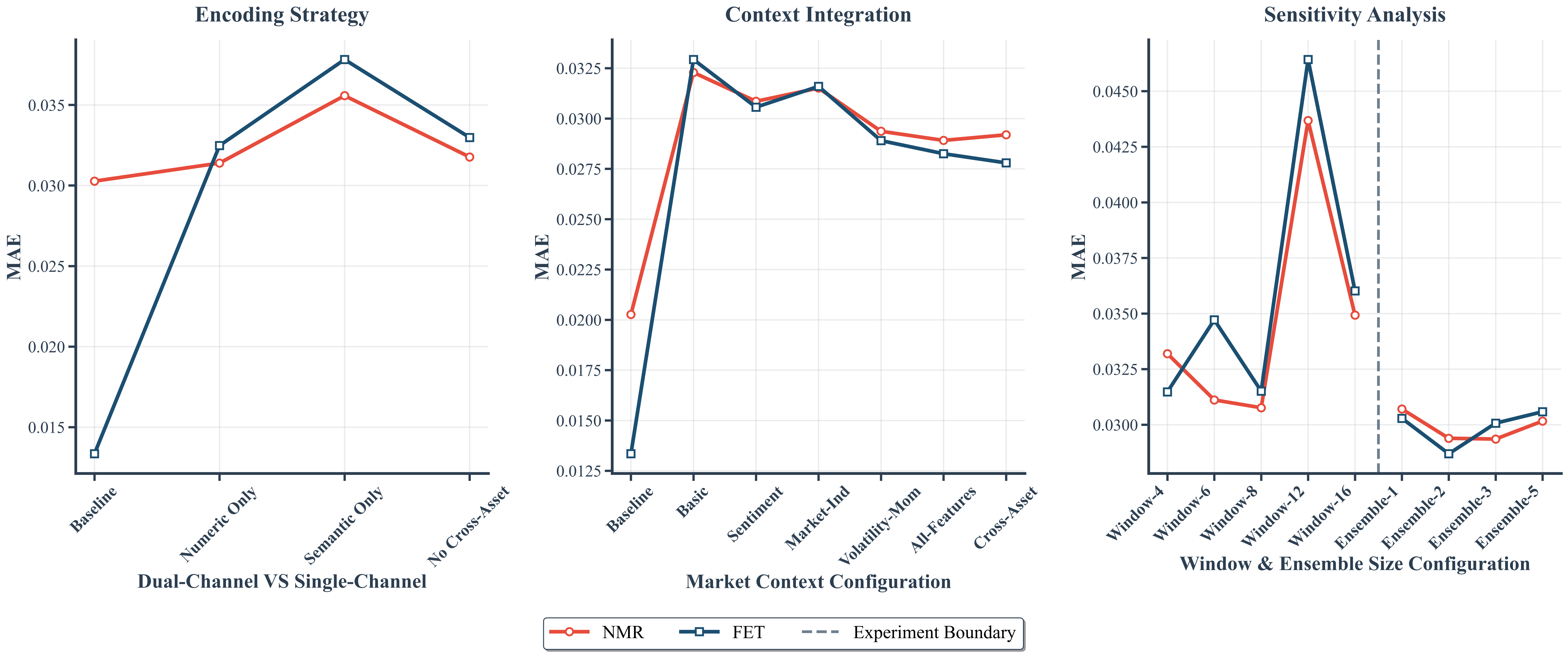}
    \caption{Impact of semantic encoding and window size configurations.}
    \label{fig:unified_ablation_study}
\end{figure}

Dual-channel semantic encoding consistently outperforms single-channel alternatives across representative cryptocurrency assets, as shown in (Figure~\ref{fig:unified_ablation_study}). The complete dual-channel implementation achieves MAE of 0.0133 for one highly volatile asset, representing 59.0\% and 64.7\% improvements over numeric-only configurations with MAE of 0.0325 and semantic-only configurations with MAE of 0.0378 respectively. A second cryptocurrency exhibits similar patterns with more modest improvements of 3.4\% over numeric-only configurations with MAE of 0.0314 and 14.9\% over semantic-only configurations with MAE of 0.0356. Cross-asset context removal produces intermediate degradation with MAE values of 0.0330 and 0.0318 for the two assets respectively, confirming the value of multi-asset information integration for capturing market interdependencies.

Market context integration experiments reveal systematic performance relationships with feature set complexity. Cross-asset expanded configurations achieve optimal MAE values of 0.0278 and 0.0292 for the two evaluated cryptocurrencies, representing 15.5\% and 9.6\% improvements over basic semantic configurations with MAE values of 0.0329 and 0.0323 respectively. Volatility-momentum feature integration provides notable contributions, achieving MAE values of 0.0289 and 0.0294 across assets. Sentiment features and market indicators demonstrate measurable but more limited performance gains compared to expanded feature sets.

Temporal window optimization identifies size 8 as optimal across evaluated assets, achieving best MAE performance of 0.0315 and 0.0308 while maintaining computational efficiency. Window size scaling beyond 8 produces systematic performance degradation, with size 12 configurations yielding MAE increases of 47.3\% and 42.0\% for FET and NMR respectively, indicating overfitting behaviors. Ensemble integration demonstrates that 2-model configurations achieve optimal performance, providing MAE improvements of 5.3\% and 4.3\% over single-model implementations. Overall, dual-channel encoding, expanded feature sets, moderate temporal windows, and ensemble integration collectively optimize ASTIF's semantic processing capabilities.

\subsection{Discussion}
\label{sec:discussion}
The experimental results address three research questions. ASTIF outperforms all baselines with MAE improvements range from 7.2\% (OCEAN) to 59.0\% (FET) across best baseline models including deep learning, and transformer architectures. Cryptocurrency assets show the largest gains, reflecting their higher volatility, regime shifts, and sensitivity to policy changes. The conditions in which ASTIF’s semantic-temporal integration provides maximal predictive advantage. The meta-learner confidence distribution (Figure~\ref{fig:meta_confidence}) confirms this interpretation. It showing high-confidence predictions when semantic and temporal signals align. Uncertainty increases during channel divergence, providing interpretable reliability estimates for risk management.

The ablation study exposes a distinct component hierarchy within ASTIF. Semantic processing emerges as the most critical component, removing it causes a 653\% increase in average MAE, while LSTM removal follows at 596\%. Uncertainty quantification and meta-learning show smaller but still substantial impacts, with 327\% and 285\% error increases, respectively. These findings confirm that every component contributes meaningfully to the ASTIF's predictive capability. Asset-level analysis reveals particularly striking dependencies: NMR and FET experience 545\% and 758\% performance deterioration when the semantic channel is removed. Such extreme sensitivity suggests that traditional numerical features alone cannot adequately capture the narrative-driven market dynamics characteristic of volatile cryptocurrency assets.

The configuration experiments establish several key findings that inform optimal framework deployment. Dual-channel encoding consistently outperforms single-channel alternatives, delivering 59.0\% and 14.9\% improvements for FET and NMR respectively. Integration of cross-asset features yields substantial gains of 15.5\% and 9.6\% over basic semantic configurations, demonstrating the value of multi-asset information in capturing market interdependencies. Window size 8 emerges as optimal across assets, effectively balancing contextual depth against overfitting risk. Notably, two-model ensembles achieve the best accuracy-efficiency balance rather than more complex configurations. These results suggest that effective cryptocurrency forecasting depends on carefully calibrated semantic-temporal integration with adaptive confidence weighting, where cross-asset correlations and policy uncertainty measures prove as critical as traditional numerical features.

\section{Conclusion}
\label{sec:conclusion}
The experimental evaluation confirms that adaptive model selection between heterogeneous predictors enables robust performance across volatile market regimes where static architectures systematically fail. ASTIF consistently outperforms established machine learning, deep learning, and transformer baselines across diverse asset classes. Performance advantages are particularly pronounced in high-volatility cryptocurrency markets characterized by regime instability and policy sensitivity. The framework validates a core theoretical premise: effective forecasting in non-stationary financial environments requires dynamic integration of semantic market understanding and temporal pattern recognition, mediated by real-time uncertainty quantification rather than fixed ensemble weights. 

Ablation experiments confirm that ASTIF's architectural components form a minimal sufficient structure. Each element, such as semantic channel, temporal modeling, uncertainty quantification, and adaptive meta-learning, contributes essential functionality. The dual-channel design enables parallel process of policy uncertainty indices and numerical price sequences through dedicated computational pathways. Configuration optimization reveals that cross-asset correlation features and moderate temporal windows maximize the accuracy-efficiency tradeoff for practical deployment. 

Practical applications span multiple stakeholder groups. Institutional investors can employ adaptive prediction strategies with interpretable confidence scores for risk-adjusted position sizing. Algorithmic trading systems can integrate regime-dependent model selection to adapt execution strategies based on market microstructure. Financial technology platforms can deploy uncertainty-aware price forecasts, communicating prediction reliability alongside point estimates. Regulatory bodies can apply semantic processing of policy uncertainty indices to assess market reactions to regulatory interventions. 

Several limitations remain, as successful prompt engineering still depends heavily on domain-specific expertise. Computational overhead constrains real-time deployment in resource-limited environments. Evaluation focuses on specific cryptocurrency and equity assets, limiting generalizability to emerging tokens or alternative asset classes. The framework processes historical data without incorporating real-time feeds or breaking news events. Future research should address real-time data integration, automated prompt optimization through reinforcement learning, multi-step forecasting capabilities, broader asset class evaluation, and edge computing optimization to enhance deployment viability.

\appendix
\label{app1}



\bibliographystyle{elsarticle-harv} 


\bibliography{references}

\end{document}